\documentclass[sigconf]{acmart}

\AtBeginDocument{%
  \providecommand\BibTeX{{%
    \normalfont B\kern-0.5em{\scshape i\kern-0.25em b}\kern-0.8em\TeX}}}

\copyrightyear{2021}
\acmYear{2021}
\setcopyright{acmcopyright}\acmConference[MM '21]{Proceedings of the 29th ACM International Conference on Multimedia}{October 20--24, 2021}{Virtual Event, China}
\acmBooktitle{Proceedings of the 29th ACM International Conference on Multimedia (MM '21), October 20--24, 2021, Virtual Event, China}
\acmPrice{15.00}
\acmDOI{10.1145/3474085.3475374}
\acmISBN{978-1-4503-8651-7/21/10}

\usepackage{graphicx}
\usepackage{amsmath}
\usepackage{subfig}
\usepackage{multirow}
\usepackage{color}

\acmSubmissionID{1124}

\begin{document}
\fancyhead{}
\title{LSTC: Boosting Atomic Action Detection with Long-Short-Term Context}

\author{Yuxi Li}
\authornote{Both authors contributed equally to this research.}
\email{yukiyxli@tencent.com}
\affiliation{%
  \institution{Tencent Youtu Lab}
  \city{Shanghai}
  \country{China}
}

\author{Boshen Zhang}
\authornotemark[1]
\email{boshenzhang@tencent.com}
\affiliation{%
  \institution{Tencent Youtu Lab}
  \city{Shanghai}
  \country{China}
}

\author{Jian Li}
\email{swordli@tencent.com}
\affiliation{%
  \institution{Tencent Youtu Lab}
  \city{Shanghai}
  \country{China}
}

\author{Yabiao Wang}
\email{caseywang@tencent.com}
\affiliation{%
  \institution{Tencent Youtu Lab}
  \city{Shanghai}
  \country{China}
}

\author{Weiyao Lin}
\authornote{Correspondence author.}
\email{wylin@sjtu.edu.cn}
\affiliation{%
  \institution{Shanghai Jiao Tong University}
  \city{Shanghai}
  \country{China}
}

\author{Chengjie Wang}
\email{jasoncjwang@tencent.com}
\affiliation{%
  \institution{Tencent Youtu Lab}
  \city{Shanghai}
  \country{China}
}

\author{Jilin Li}
\email{jerolinli@tencent.com}
\affiliation{%
  \institution{Tencent Youtu Lab}
  \city{Shanghai}
  \country{China}
}

\author{Feiyue Huang}
\email{garyhuang@tencent.com}
\affiliation{%
  \institution{Tencent Youtu Lab}
  \city{Shanghai}
  \country{China}
}


\begin{abstract}
  In this paper, we place the atomic action detection problem into a Long-Short Term Context (LSTC) to analyze how the temporal reliance among video signals affect the action detection results. To do this, we decompose the action recognition pipeline into short-term and long-term reliance, in terms of the hypothesis that the two kinds of context are conditionally independent given the objective action instance. Within our design, a local aggregation branch is utilized to gather dense and informative short-term cues, while a high order long-term inference branch is designed to reason the objective action class from high-order interaction between actor and other person or person pairs. Both branches independently predict the context-specific actions and the results are merged in the end. We demonstrate that both temporal grains are beneficial to atomic action recognition. On the mainstream benchmarks of atomic action detection, our design can bring significant performance gain from the existing state-of-the-art pipeline. The code of this project can be found at \textcolor{blue}{\url{https://github.com/TencentYoutuResearch/ActionDetection-LSTC}}
\end{abstract}




\maketitle

\section{Introduction}

\begin{figure}
    \centering
    \includegraphics[width=0.45\textwidth]{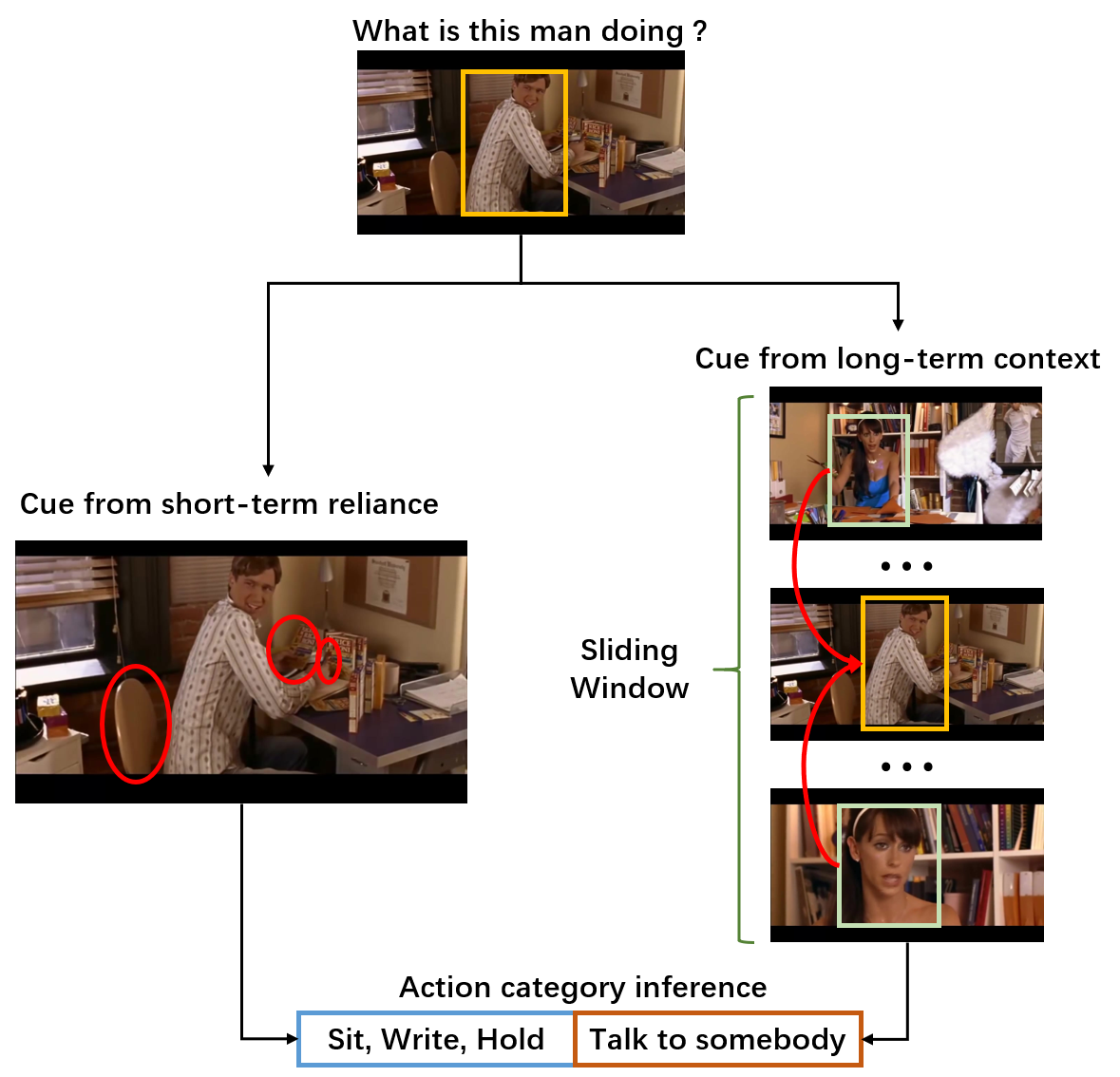}
    \caption{A example of independent temporal inference of different scope. The left branch shows short-term reliance on surrounding. The right branch illustrates long-term cues from action subjects contained by other clips.}
    \label{fig:concept}
\end{figure}

It has been widely studied on how to correctly recognize the human actions from videos, most of the previous efforts~\cite{C3D,I3D,SlowFast,TwoStream} resort to temporal information modeling since in some traditional benchmarks~\cite{UCF101,I3D}, different action instances show distinct visual motion patterns across time. However, things are different when it comes to the \textbf{atomic} action problem~\cite{AVA,HIEVE}. Atomic action detection aims at localizing persons of extremely subtle behaviors (e.g. \emph{staring at something} or \emph{calling the phone}), where the visual motion information is limited. Under this scenario, visual context plays crucial role because most action instances involve different kinds of interactions, either among actors or between human and objects. Therefore, it is a key factor towards high-quality detection to capture contextual interaction cues in spatiotemporal domain.

There are some attempts to incorporate the spatiotemporal interaction with atomic action detection pipelines~\cite{LFB, girdhar2019actionTransformer, wu2020context}. Nevertheless, they ignore the discrepancy between different contextual interactions. Within a short temporal scope, the subject of an action can interact with any space around it, either certain background objects or subjects of another action instance. On the other hand, with a longer temporal distance, the subjects of different action instances are more likely to interact with other subjects, in this case, the background or objects information is less informative to help reason the action types. Figure~\ref{fig:concept} shows an example of such inference process relying on both scopes of context. In this example, some action types are easy to be inferred at a glance of the short clip (e.g. \emph{sitting} and \emph{writing}), since the background provides cues to support the judgment (the chairs and pen). However, we can not reason that the man is talking with others if not inquiring the action instances from video clips before and after the current frame. This observation indicates that both short-term and long-term context are crucial to completely recognize the action of a person under complex interaction scenarios.
Further, interactions within different time scope are usually independent of each other and can provide complementary cues for the recognition of current action. In the example illustrated in Figure~\ref{fig:concept}, we can see the short-term cue and long-term counterpart are helpful to infer distinct sets of action types, the ignorance of one kind of context cue does not affect the reasoning from other cue. This indicates that context-based atomic action detection can be decoupled into independent inference process of different temporal scope.

With the inspiration above, we build a context-based probability graph model to analyze the \textbf{reliance} and \textbf{independence} between different variables in atomic action detection problem. Guided by the analysis, we design an atomic action detection pipeline with independent long-short-term reasoning procedure. In short-term reasoning process, we try to aggregate local information from dense context representation with a pixel-wise aggregation mechanism since the action subjects can interact with any specific spatial or temporal fragment within a short time slot. On the other hand, in a longer temporal scope, a discrete actor-wise feature selection and refinement mechanism is designed to gather informative context from a long-term feature bank~\cite{LFB}. Different from the long-term operation in previous works~\cite{LFB,wu2020context} which only models person-to-person interaction and refered as a first-order attention model, we resort to a decoupled second order attention module to exploit person-to-pair relationship with feasible complexity. In this way we ensure that the long-term inference process can independently predict specific class given the context and we demonstrate that such design outperforms the feature-level operation in~\cite{LFB}. We conduct experiments on two benchmarks focusing on atomic action recognition, AVA~\cite{AVA} and HiEve~\cite{HIEVE}, demonstrating that our two-scope inference pipeline can bring significant improvement over existing state-of-the-art methods. In a nutshell, the contribution of this paper can be summarized as:

\begin{itemize}
    \item The atomic action problem is decoupled as a long-short-term inference task and we propose a parallel reasoning pipeline to solve the problem in both temporal grain. 
    \item A local context aggregation branch is designed to capture helpful information from dense spatiotemporal feature within a short-term scope.
    \item A high-order long-term attention modeling process is incorporated with our framework to boost its actor-wise long-term reasoning ability.
\end{itemize}

\section{Related Works} \label{Related Works}

\subsection{Action recognition}
Deep learning technique has pushed the advance of video action recognition to a large extent. Recent works have been trying to design effective CNN architectures for action recognition in short clips~\cite{C3D,I3D,S3D,SlowFast,TwoStream,TSN}. Two-stream ConvNets~\cite{TwoStream,TSN} are designed with spatial and temporal branches to capture the complementary appearance information from still frames and visual motion patterns. 3D-CNN~\cite{C3D,I3D,SlowFast} directly model the spatial and temporal information with 3D convolution kernels by inflating the ImageNet~\cite{ImageNet} pre-trained model and training on input clip of different sampling rate. In addition, there are lines of works resorting to the correlation mechanism at different feature dimension and scale to encode high-level relationship features~\cite{Nonlocal,Wang_2020_CVPR} for accurate action classification. Nevertheless, most of the existing works mentioned above focus on properly capturing the motion within a short temporal span and are more suitable for normal action recognition than subtle atomic-level recognition.

\subsection{Atomic action detection} Atomic action detection aims at localizing and recognizing subtle human actions simultaneously in video clips, where the visual motion is not salient and the action subjects are involved in different types of interactions. A large-scale dataset AVA~\cite{AVA} is constructed focusing on solving such challenging problem, accompanied with an actor-centric two-stream model as the baseline. Some recent works follow the object detection paradigm by first localizing person bounding box with pre-trained human detector~\cite{FasterRCNN} and utilizing the ROI features from deep features to predict action class~\cite{SlowFast,ACRN}. Besides, a short-term graph model~\cite{zhang2019Graph} is built upon the spatiotemporal location of objects to enhance the performance on complex action discrimination. Finally, there are also some works applying the attention mechanism~\cite{vaswani2017AttentionAllYouNeed} to capture either short-term dense interaction~\cite{girdhar2019actionTransformer} or discrete long-term relation~\cite{LFB}. 
However, they solely consider relation under a single time scope, while our work take both long-term and short-term interaction into account and elaborately process them at different level.

\subsection{Long-term context reasoning}
 Context information plays an important role in highly self-correlated data, in most cases the semantics can be directly inferred from the contextual cues. This property has been widely studied in the field language processing,~\cite{language_model} first proposed a nueral language model to model the concurrent likelihood of certain objective words and their surrounding text. The BERT model~\cite{devlin-etal-2019-bert} further combined this property with attention-based selection scheme for the self-training of language model. There are also some trials in vision problems to incorporate contextual modeling with action recognition~\cite{yao2010modeling, marszalek2009actionsInContext,girdhar2017attentionalPooling,sharma2015actionVisualAtt}, but mostly focus on the spatial and short-term context, while accurate atomic action detection requires both short-term and long-term cues in spatiotemporal domain. 

\section{Long-Short-Term Context}

\subsection{A probability view of context information}
\begin{figure}
    \centering
    \subfloat[Probability graph of a normal action recognition problem.]{\includegraphics[width=0.4\textwidth]{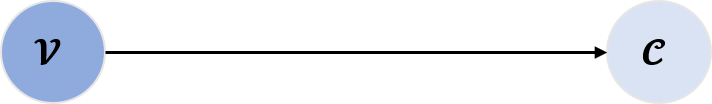}
    \label{fig:prob0}}\\
   \subfloat[The probability graph of long-short-term modeling in atomic action detection.]{\includegraphics[width=0.4\textwidth]{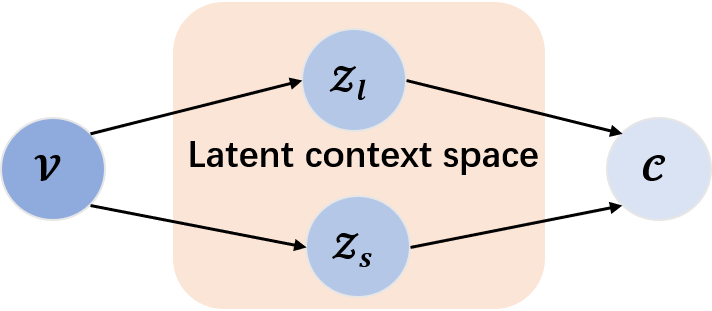}
    \label{fig:prob}}
    \caption{Probability graph model of previous action methods and our long-short-term analysis.}
\end{figure}

Before introducing our framework, we first give a probability formulation of atomic action detection problem. By our definition, $\mathcal{V}$ is the actor-wise representation of an input short video clip and $\mathcal{C}$ is the variable indicating the action categories contained by the clip. In traditional action recognition paradigm, the distribution of action class is directly dependent on input clips as $\mathcal{P}(\mathcal{C}|\mathcal{V})$ (illustrated as Figure~\ref{fig:prob0}). Nevertheless, we analyze the problem with a more complex model with the hypothesis that the action type distribution is not directly conditioned on $\mathcal{V}$ but indirectly related to it via two intermediate latent variables $\mathcal{Z}_l$ and $\mathcal{Z}_s$, where $\mathcal{Z}_l$ denotes the long-term contextual information and $\mathcal{Z}_s$ is short-term reliance variable. Both latent variables are conditioned on the input clip and together determine the distribution of action class. Figure~\ref{fig:prob} depicts the probability graph of the reliance between variables defined above.

Since the action type $\mathcal{C}$ is not directly related to the input videos, we make attempt to model the joint distribution of tuple $(\mathcal{Z}_l, \mathcal{Z}_s, \mathcal{C})$ conditioned on input as $\mathcal{P}(\mathcal{Z}_l, \mathcal{Z}_s, \mathcal{C}|
\mathcal{V})$. From Figure~\ref{fig:prob}, it is easy to draw following conditional independent relationship
\begin{equation}
\mathcal{C}\perp\mathcal{V}|(\mathcal{Z}_l, \mathcal{Z}_s) \quad \mathcal{Z}_l\perp\mathcal{Z}_s|\mathcal{V}
\end{equation}
Hence the joint probability distribution can be decoupled as
\begin{equation}\label{eq:prob}
\begin{split}
    \mathcal{P}(\mathcal{Z}_l, \mathcal{Z}_s, \mathcal{C}|\mathcal{V}) = & \mathcal{P}(\mathcal{Z}_l, \mathcal{Z}_s|\mathcal{V})\mathcal{P}(\mathcal{C}|\mathcal{Z}_l, \mathcal{Z}_s, \mathcal{V}) \\
     = & \mathcal{P}(\mathcal{Z}_l, \mathcal{Z}_s|\mathcal{V})\mathcal{P}(\mathcal{C}|\mathcal{Z}_l, \mathcal{Z}_s) \\ 
     = & \mathcal{P}(\mathcal{Z}_l|\mathcal{V})\mathcal{P}( \mathcal{Z}_s|\mathcal{V})\mathcal{P}(\mathcal{C}|\mathcal{Z}_l, \mathcal{Z}_s)
\end{split}
\end{equation}
In Equation~(\ref{eq:prob}) we decouple the joint probability into three terms. $\mathcal{P}(\mathcal{Z}_l|\mathcal{V})$ and $\mathcal{P}(\mathcal{Z}_s|\mathcal{V})$ respectively model the long-term and short-term reliance between video clips and their context, while $\mathcal{P}(\mathcal{C}|\mathcal{Z}_l, \mathcal{Z}_s)$ can be regarded as a joint discrimination function to determine the action class according to long-short-term context.

\begin{figure*}
    \centering
    \includegraphics[width=0.9\textwidth]{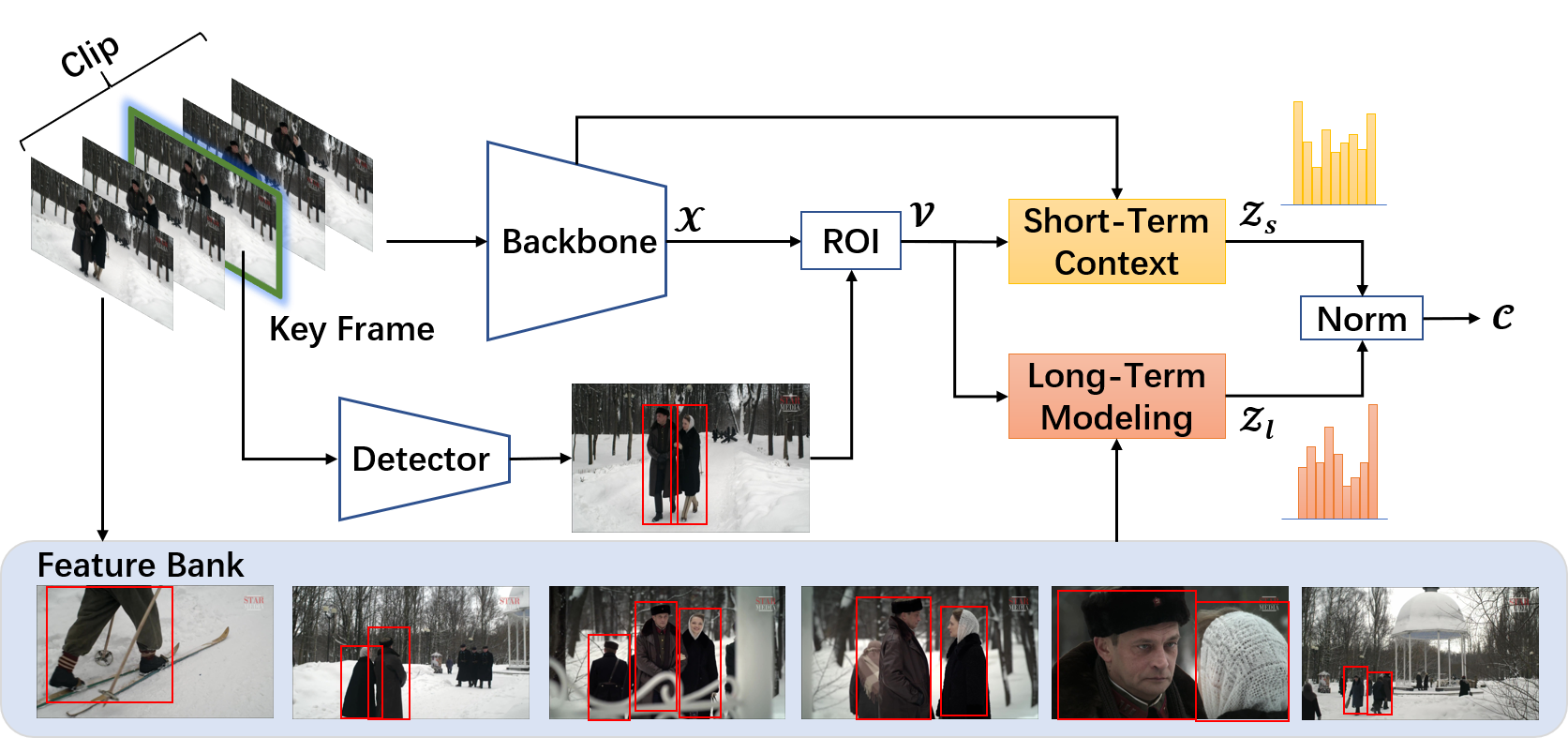}
    \caption{Overview of our pipeline with long-term and short-term context module.}
    \label{fig:pipeline}
\end{figure*}

With the discussion above, we design our pipeline with a paradigm of parallel processing and late fusion. The framework is illustrated in Figure~\ref{fig:pipeline}. In this figure, we utilize the 3D deep neural network to extract feature and take the deep features pooled from detected actor boxes as our actor-wise representation $\mathcal{V}$, where detected actor boxes are obtained via a person detector~\cite{FasterRCNN} applied on the center frame of input clip (noted as the ``key frame'' in the rest of this paper). Next one short-term context branch is designed to aggregate helpful short-term features from the dense spatiotemporal features with the guidance of $\mathcal{V}$. Meanwhile, a long-term context processing branch attempts to mine high-level actor-wise interaction with $\mathcal{V}$ from a longer temporal scope. Finally, the output context $\mathcal{Z}_l$ and $\mathcal{Z}_s$ are fused together to determine final action category $\mathcal{C}$. More detail will be introduced in following sections.

\subsection{Short-term local aggregation}

Although the person-wise deep feature from ROI-pooling operation~\cite{FasterRCNN} can extract helpful information for action detection, more accurate atomic action recognition relies on the surrounding cues of actors. Therefore, we make attempt to aggregate local spatiotemporal information from input clip to build more descriminative short-term action type representation $\mathcal{Z}_s$, which is conditioned on $\mathcal{V}$.

To be specific, the output 3D feature from backbone network is denoted as $\mathcal{X} \in \mathcal{R}^{HWT \times d}$, where $H,W,T$ are the spatial and temporal dimension of the clip after downsampling in the backbone and $d$ is the feature dimension.   Suppose the output tensor $\mathcal{V}$ is of size $N \times d$ after the ROI operation, where $N$ is the number of detected actors from the key frame, we take this actor-centric feature as query to generate a spatiotemporal attention map from $\mathcal{X}$ as

\begin{equation}
    \mathcal{A}(\mathcal{X}, \mathcal{V}) = \textit{softmax}\left(\mathcal{V}W_{\mathcal{A}}\mathcal{X}^T\right)
\end{equation}
where $\mathcal{A}(\mathcal{X}, \mathcal{V}) \in \mathcal{R}^{N\times HTW}$ indicates the dense reliance between actors and its spatiotemporal surrounding and $W_{\mathcal{A}} \in \mathcal{R}^{d\times d}$ is a learnable matrix to indicate the importance of correlation between each dimension pair. The softmax operation is applied as an normalization function over the spatiotemporal dimension of original 3D features. With this attention map, we can aggregate the spatiotemporal context from surrounding feature $\mathcal{X}$ as a person-guided feature $\mathcal{V}_s$

\begin{equation}
    \mathcal{V}_s = \mathcal{A}(\mathcal{X}, \mathcal{V})\phi(\mathcal{X};W_{\mathcal{V}})
\end{equation}
where $\phi(\cdot)$ is a simple linear mapping function parameterized by $W_{\mathcal{V}}$. Since both the actor-centric feature $\mathcal{V}$ and its corresponding surrounding descriptor $\mathcal{V}_s$ encode the action-specific short-term context, we combine them together to get our final short-term local aggregation output as

\begin{equation}
    \mathcal{Z}_s = h(\textit{FFN}([\mathcal{V}, \mathcal{V}_s]; W_f); W_s)
\end{equation}
where $[\cdot,\cdot]$ denotes the concatenation operation along the channel dimension, \textit{FFN} is short for feed forward network, which is composed of a multi-layer perceptron and $W_f$ is its learnable parameters. Finally the merged feature is processed by a short-term discriminator function $h(\cdot)$ parameterized by $W_s$ and mapped to the short-term context latent space $\mathcal{Z}_s \in \mathcal{R}^c$, where $c$ is the number of classes.

\subsection{High order long-term interaction modeling}

\begin{figure}
    \centering
    \includegraphics[width=0.46\textwidth]{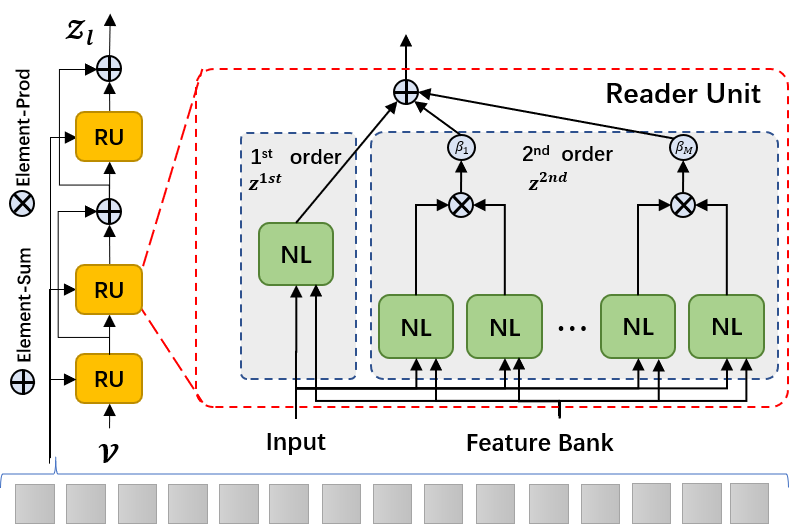}
    \caption{Detailed structure of long-term modeling process. Where ``RU'' is short for the reader unit and ``NL'' is short for NonLocal block.}
    \label{fig:reader}
\end{figure}

 In addition to the surrounding objects within the short input clip, the long-term interaction with other actors from different video shot can also provide helpful cues for action detection in current clip. To modeling such long-term reliance, attention-based approach is widely used in previous works focusing on video analysis~\cite{LFB,STM,wu2020context}, in~\cite{Nonlocal} such attention mechanism is summarized as Equation~(\ref{eq:1-order}) and implemented with a NonLocal block~\cite{Nonlocal}.
 \begin{equation}\label{eq:1-order}
     z^{1st}_i = \frac{1}{\mathcal{T}(v_i)}\sum_{j\in \mathcal{C}_i}s\left(v_i, v_j\right)g(v_j)
 \end{equation}
where $\mathcal{C}_i$ is the context space of data sample $v_i$, $s(\cdot,\cdot)$ is a pair-wise similarity metric, $g(\cdot)$ is a mapping function, and $\mathcal{T}(v_i)=\sum_{j\in \mathcal{C}_i}{s\left(v_i, v_j\right)}$ is normalization term. Equation~(\ref{eq:1-order}) can be refered as a \textbf{first-order} attention model since it resorts to the correlation to single instance in context space. However, in a more comprehensive scope, the semantic reliance of an object in videos can be not only relative to pairwise relationship, in contrast, the co-occurrence of other instances in context space can also provide important cues for semantic inference~\cite{pennington2014glove}. Therefore in our design, we try to introduce a more representative \textbf{second-order} attention model to exploit the correlation between single-person and co-occurrence pairs
 \begin{equation}\label{eq:2-order}
     z^{2nd}_i = \frac{1}{\mathcal{T}'(x_i)}\sum_{(j,k)\in \mathcal{C}_i\times \mathcal{C}_i}s'\left(v_i, v_j, v_k\right)g'(v_j, v_k)
 \end{equation}
 where $\mathcal{T}'(v_i)=\sum_{(j,k)\in \mathcal{C}_i\times \mathcal{C}_i}s'\left(v_i, v_j, v_k\right)$, however, calculating the second-order attention in Equation~(\ref{eq:2-order}) results $\mathcal{O}(|\mathcal{C}_i|^2)$ complexity for each person, which is infeasible when context space is large. Inspired by the attempts in tensor decoupling~\cite{guo2018nd}, we approximate such attention with decoupling to achieve $\mathcal{O}(|\mathcal{C}_i|)$ complexity
 \begin{equation}\label{eq:decouple}
\begin{split}
    z^{2nd}_i & \approx \frac{1}{\mathcal{T}'(x_i)}{\sum_{(j,k)\in \mathcal{C}_i\times \mathcal{C}_i}s'_{1}\left(v_i, v_j\right)s'_{2}\left(v_i, v_k\right)g'_{1}(v_j)g'_{2}(v_k)} \\
     & = {z^{1st}_{i,1}}{z^{1st}_{i,2}}\\
    z^{1st}_{i,l} & = \frac{1}{\sum_j{s'_{l}(v_i, v_j)}}\sum_{j\in \mathcal{C}_i}s'_{l}\left(v_i, v_j\right)g'_{l}(v_j) \quad l = 1,2
\end{split}
\end{equation}
Via Equation~(\ref{eq:decouple}), we decouple the second-order attention into the form of multiplication between output of two first-order NonLocal blocks. Combining the attention in different order $z^{1st}_i,z^{2nd}_i$ we can build our long-term interaction module as Figure~\ref{fig:reader}, where we follow the experience of multi-head and cascaded attention~\cite{vaswani2017AttentionAllYouNeed}. To be specific, sequentially we cascade $K$ Reader Units (RU) to recurrently extract and refine long-term reliance to get output long-term representation $\mathcal{Z}_l$. Within each Reader Unit, we apply $M$ parallel NonLocal pairs and utilize learnable weight parameters $\beta_m$ to aggregate the results to form approximate second-order attention. In implementation, we take the long-term feature bank~\cite{LFB} as the context space for each person from the input clip.

\subsection{Joint discrimination}

In the design above, either $\mathcal{Z}_s$ or $\mathcal{Z}_l$ can be regarded as action estimation in the latent space, while in a global sense, the action category of specific person is determined by both long-term and short-term context, therefore we simply incorporate these two types of context of two temporal scope with a late-fusion mechanism

\begin{equation}
    \mathcal{C} = \textit{Norm}(\mathcal{Z}_l + \mathcal{Z}_s)
\end{equation}
where \textit{Norm} is a normalization function to map the context variable to class probability space $[0,1]^{c}$. During the training stage, the output distribution is supervised by the given action labels via  cross entropy loss function.

\section{Experiments}


\label{Experiments}

\subsection{Experimental settings}
\label{Experimente:1}

\subsubsection{Datasets}
\label{Experimente:1.1}

\textbf{- Atomic Visual Action (AVA)}~\cite{AVA} is the first dataset constructed for spatiotemporal detection of subtle atomic actions including $80$ categories of actions in total. This dataset consists of $235$ long movie sequences for training and other $64$ for validation. The videos are annotated with boxes on each frame at a frequency of $1$FPS, each actor is associated with one or more action labels. 

\noindent\textbf{- Human in Events (HiEve)}~\cite{HIEVE} is a recently public benchmark towards comprehensive analysis on surveillance video in events. Different from AVA, it contains more crowded daily and emergency scenarios. The dataset includes $14$ different action classes and consists of 32 video sequences ($19$ for training and other $13$ for testing). Each sequence last for about 2 minutes on average, these videos are annotated with bounding boxes and action labels every 20 frames.

\subsubsection{Evaluation protocol}
\label{Experimente:1.2}

\noindent\textbf{- Frame-mAP}. For both benchmark, we report the frame-level mAP value as the object detection tasks~\cite{VOC,COCO} for performance evaluation, if an action proposal has overlap larger than $\delta$ with a ground-truth box and has the same action label, it is regarded as positive proposal. In AVA, $\delta$ is set as $0.5$, following~\cite{AVA}, we only evaluate the $60$ most common action classes. In HiEve, we follow~\cite{HIEVE} to compute mAP score under $\delta=[0.5, 0.6, 0.75]$ and calculate the average score as final performance. Besides, we also report weighted-mAP in HiEve, which focuses more on complex and crowd scenes in videos and assign larger evaluation weight to frame with large crowd index. 
\begin{figure}
    \centering
    \includegraphics[width=0.475\textwidth]{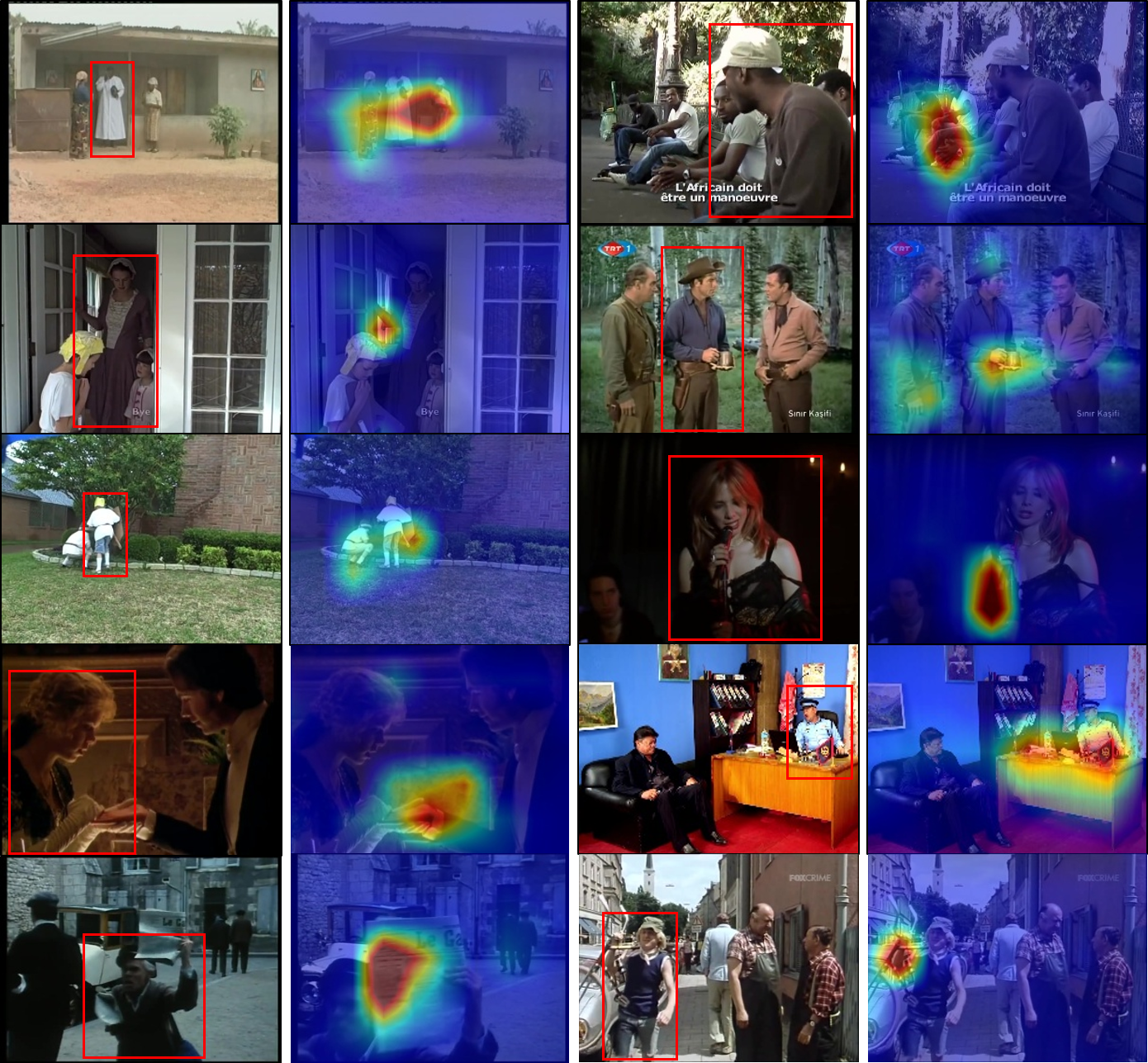}
    \caption{Attention heatmap from short-context module. The left column indicates the detected person in red box, the right column shows its corresponding spatiotemporal attention. (best viewed in color)}
    \label{fig:heat}
\end{figure}

\subsubsection{Implementation detail}
\label{Experimente:1.3}

\noindent\textbf{- Person detector.} For AVA, We apply the person detector from~\cite{LFB} to detect actors on key frames, which is a modified version of FPN-Faster-RCNN~\cite{FasterRCNN,lin2017feature} and pretrained on COCO Keypoints~\cite{COCO}, the detector is then finetuned on AVA. For HiEve, we directly use the public detected results from the multi-object tracking track of HiEve challenge\footnote{http://humaninevents.org}.

\noindent\textbf{- Baseline model.} In our experiments, we take the SlowFast~\cite{SlowFast} network with ResNet~\cite{he2016deep} backbone and Inflated-3D network (I3D)~\cite{Nonlocal} as our base network for spatiotemporal feature extraction. All these base model in our experiments are first pretrained on the Kinetics dataset~\cite{I3D} and then finetuned on the target benchmark. In our implementation, we append an additional $1 \times 1 \times 1$ convolution after the network trunk to reduce the channel size by half. For the ROI operation, we separately apply average and max pooling along the temporal dimension of the output 3D feature to get two 2D spatial feature, then take ROIAlign~\cite{He_2017_ICCV} on each spatial feature, the pooled feature vectors are summed together as person-specific feature. In a baseline setting, this feature is directly sent to an action classifier. 

\noindent\textbf{- Training detail.} We implement our framework with the PySlowFast platform~\cite{PySlowfast}. We select each annotated frame as the key frame, and uniformly sample $32$ frames centered on this frame with a sampling rate of $2\times$, this results in a short input clip with temporal endurance of around $2$ seconds. The training process is splitted into two stage, first we train a base model without long-term context modeling, and extract the ROI feature into long-term feature bank, then in the second stage, we train the network with both long-term and short-term context. During training, we take both the annotated bounding boxes and detected person boxes with confidence score larger than $0.9$ as box proposals to extract person-wise ROI feature to bridge to localization gap between person detector and groundtruth. The network is trained on 8 GPUs with a total batch size of $64$ clips. The learning rate is initialized with a warmup value of $1.25\times 10^{-4}$ and linearly increases to $0.1$ in $5$ epochs, then the learning rate is degraded by a factor of $10$ every $5$ epochs. The network parameters are optimized by SGD algorithm with a weight decay ratio of $10^{-7}$. Similar to~\cite{SlowFast}, we apply random flip and crop to augment the data.


\subsection{Ablation studies}
\label{Experimente:2}

We conduct our ablation studies on AVA to analysis the effect of each component in our framework. If not specified, we take v2.2 for both training and evaluation in default.

\begin{table}
\setlength{\belowcaptionskip}{-0.cm}
    \centering
    \begin{tabular}{c|cc|c}\hline
        \multirow{2}{*}{backbone} & \multicolumn{2}{|c|}{component} & \multirow{2}{*}{mAP@$0.5$} \\\cline{2-3}
         & $\mathcal{Z}_s$ & $\mathcal{Z}_l$ & \\ \hline 
          \multirow{3}{*}{I3D} & & & 15.8 \\
            & \checkmark & & 20.7 \\
            & \checkmark & \checkmark & \textbf{22.4} \\\hline
            \multirow{3}{*}{Res50} & & & 24.6 \\
            & \checkmark & & 26.1 \\
            & \checkmark & \checkmark & \textbf{28.7} \\\hline
             \multirow{3}{*}{Res101-NL} & & & 27.2 \\
            & \checkmark & & 27.7 \\
             & \checkmark & \checkmark & \textbf{30.3} \\\hline
    \end{tabular}
    \caption{Ablation study on the directly effect of long-short-term context on AVA v2.2 validation set.}
    \label{tab:component}
\end{table}

\textbf{(1). How does each context component directly affect the final results ?} To investigate the effectiveness of long-term and short-term context, we conduct experiments with baseline models of different backbones. We start from a baseline model with a simple ROI operation followed by a classier, and then test on our trained model from the first training stage without long-term context, finally report results on full model. The corresponding performance is listed in Table~\ref{tab:component}. From the table we observe that both long-term interaction modeling and short-term local aggregation can bring improvement in overall detection score regardless of the backbone we used, where our full model can achieves at most \textbf{+6.6} higher performance than the simple baseline. We find that the attention-based local aggregation is less helpful on the SlowFast with Res101-NL backbone than I3D and SlowFast-Res50, this can be due to the fact that the inserted NonLocal block already aggregate some global and surrounding information from the clip in the feature extraction stage. In contrast, we find the short-term improvment is more salient when the backbone is weaker.

\begin{figure*}
    \centering
    \includegraphics[width=0.95\textwidth]{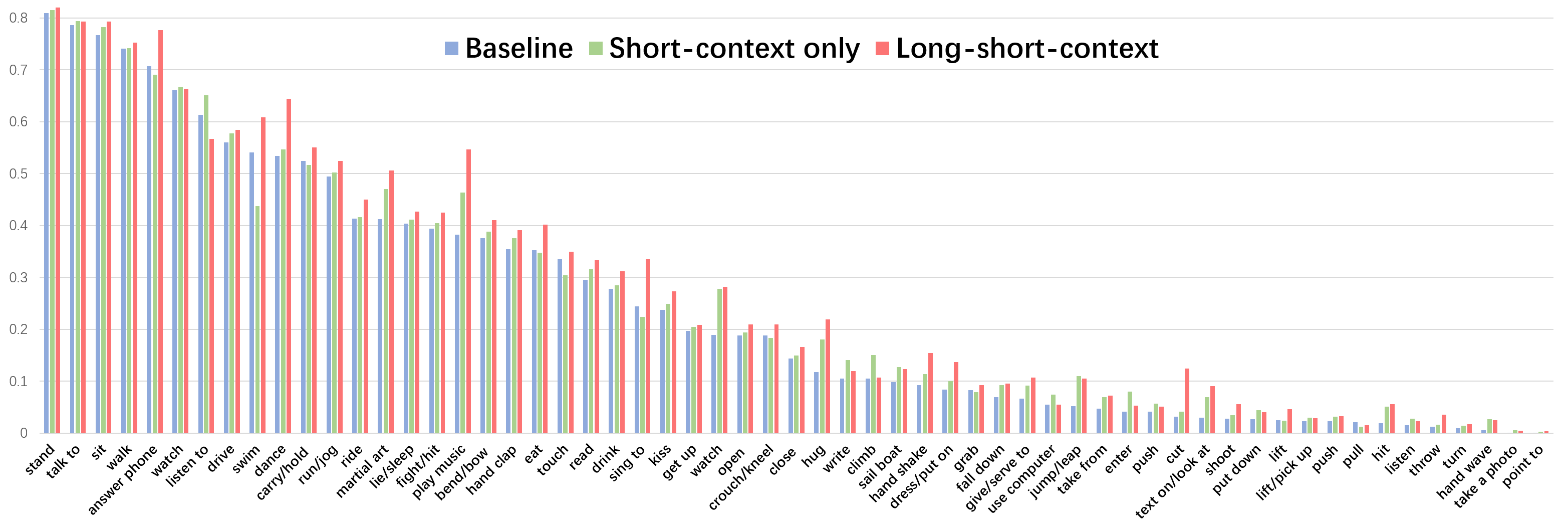}
    \caption{AP@0.5 value for each action class on AVA v2.2 validation set in the descending order of baseline performance, obtained from models with SlowFast network and Res50 backbone. ( best viewed in color)}
    \label{fig:per_class}
\end{figure*}

\begin{table}[]
\setlength{\abovecaptionskip}{0.cm}
\setlength{\belowcaptionskip}{-0.cm}
    \centering
    \begin{tabular}{cc|c|c}\hline
        \multicolumn{2}{c|}{context source} & \multirow{2}{*}{mAP@0.5} & \multirow{2}{*}{Increment} \\\cline{1-2}
         short-term & long-term & & \\\hline
         Res50 & Res50 & 28.7 & - \\
         Res50 & Res101-NL & 30.0 & + 1.3 \\
         Res101-NL & Res50 & 29.4 & + 0.7 \\
         Res101-NL & Res101-NL & \textbf{30.3} & \textbf{+ 1.6} \\\hline
    \end{tabular}
    \caption{Investigation on different combination of long-short-term context sources.}
    \label{tab:source}
\end{table}
\begin{figure}
    \centering
    \includegraphics[width=0.45\textwidth]{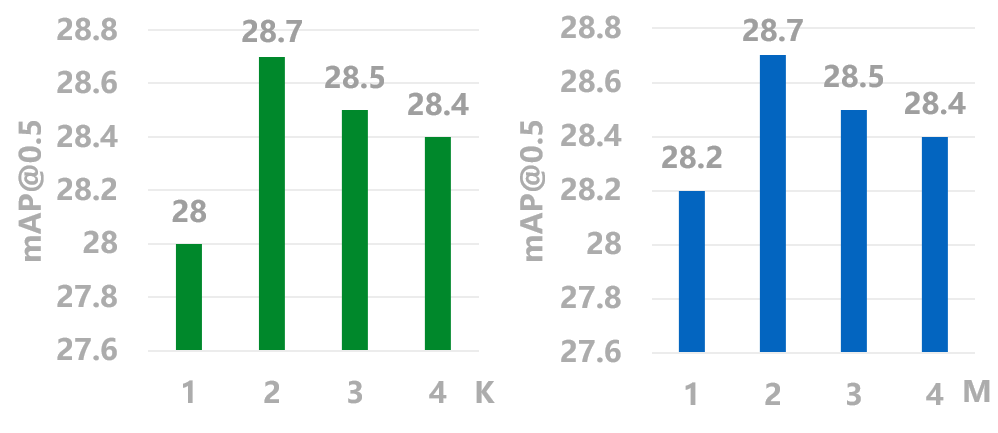}
    \caption{Invstigation on hyperparameter $K$ and $M$.}
    \label{fig:reader-results}
\end{figure}
In Figure~\ref{fig:per_class}, we visualize the per-class AP value in terms of the output from SlowFast-Res50 backbone with different settings. We find the short-term context is more helpful to recognize actions with semantics highly related to the scene or surrounding objects (e.g. \emph{listen to} and \emph{climp}). On the other hand, the long-term interaction modeling is benifical especially for actions involving group activities or with long endurance (e.g. \emph{dance} and \emph{swim}). Besides, we visualize the person-guided spatiotemporal attention map in Figure~\ref{fig:heat}, we can see, our short-context aggregator can properly catch some critical part from the short-term clip for corresponding action recognition.

\textbf{(2). How does the parameter $K$ and $M$ affect the performance ?} Next we investigate the optimal configuration ($K$ for Reader Unit number and $M$ for number of second attention head) for long-term modeling with the SlowFast-res50 backbone. The results are reported in Figure~\ref{fig:reader-results}. We see that both parameters achieves optimal results at the value of $2$, when $K \& M$ increase, the results are degraded. This is probably due to more complicated model structure harmful to model training.

\textbf{(3). What if the long and short context are from different backbones ?} In our normal setting, the long-short-term context is extracted from the same backbone, in Table~\ref{tab:source}, we investigate the effect of different context source combination. In detail, we adopt the Res50 and Res101-NL as base models, taking one for long-term feature bank extraction and the other as short-term backbone to process input video clip. The results are compared with our full model with either backbone. From Table~\ref{tab:source} we observe that when the quality of either context is improved, there will be obvious enhancement in recognition results. Further, we see improving long-term context brings more obvious benefits ($+1.6$), demonstrating that our designed high-order attention can effectively harness the long-term reliance.


\textbf{(4). Comparison with the Feature Bank Operator (FBO)~\cite{LFB}}. Some other methods to deal with long-term features are proposed, within which the FBO~\cite{LFB} is the most similar counterpart to ours, which attempts to fuse the long-term information in feature representation with a first-order attention model. We compare this operation with our high-order modeling part on different backbones. To do this, we take the backbone with pure short-term context as a zero-order model, and append first-order (FBO) and second-order (LSTC) attention respectively. In Table~\ref{tab:FBO} we list the comparison results on Res50 and Res101-NL backbone, we observe that either backbone benefit more from our long-term modeling mechanism than FBO, this indicates that our scheme can capture the long-term reliance more sufficie ntly.

\begin{table}[]
    \centering
    \begin{tabular}{c|c|c|c}\hline
        backbone & setting & long-term order & mAP@0.5 \\\hline
        \multirow{3}{*}{Res50} & short-term only & zero order & 26.1 \\
        & + FBO~\cite{LFB} & first order & 27.5 \\
        & + LSTC (ours) & second order & \textbf{28.7} \\ \hline
        \multirow{3}{*}{Res101-NL} & short-term only & zero order & 27.7 \\
        & + FBO~\cite{LFB} & first order & 29.4 \\
        & + LSTC (ours) & second order &\textbf{30.3} \\ \hline
    \end{tabular}
    \caption{Comparison results between our long-term modeling and FBO~\cite{LFB} on AVA v2.2 validation set.}
    \label{tab:FBO}
\end{table}

\subsection{Comparison with state-of-the-art methods}
\label{Experimente:3}

\begin{table}[]
    \centering
    \small
    \begin{tabular}{c|c|c|c}\hline
        method & backbone & pretrain & mAP@0.5 \\\hline
         Baseline~\cite{AVA} & I3D & Kinetics-400 & 15.6 \\
         ACRN~\cite{ACRN} & S3D & Kinetics-400 & 17.4 \\
         VAT~\cite{girdhar2019actionTransformer} & I3D & Kinetics-400 & 25.0 \\
         SMAD~\cite{zhang2019Graph} & I3D & Kinetics-400 & 22.2 \\
         SlowFast~\cite{SlowFast} & Res50 & Kinetics-400 & 24.2 \\
         SlowFast~\cite{SlowFast} & Res101-NL & Kinetics-600 & 27.3 \\
         LFB~\cite{LFB} & Res50-NL & Kinetics-400 & 25.8 \\
         LFB~\cite{LFB} & Res101-NL & Kinetics-400 & 27.7 \\
         C-RCNN~\cite{wu2020context} & Res50-NL & Kinetics-400 & 28.0 \\ \hline
         LSTC (ours) & Res50 & Kinetics-400 & 28.4 \\
          LSTC (ours) & Res101-NL & Kinetics-600 & \textbf{30.0} \\\hline
    \end{tabular}
    \caption{Comparison results with other methods on AVA v2.1 validation set.}
    \label{tab:ava_v21}
\end{table}

\begin{table}[]
    \centering
    \small
    \begin{tabular}{c|c|c|c}\hline
        method & backbone & pretrain & mAP@0.5 \\\hline
         SlowFast~\cite{SlowFast} & Res50 & Kinetics-400 & 24.9 \\
         SlowFast~\cite{SlowFast} & Res101-NL & Kinetics-600 & 29.2 \\
         AVSF~\cite{xiao2020audiovisual} & Res50 & Kinetics-400 & 25.9 \\
         AVSF~\cite{xiao2020audiovisual} & Res101-NL & Kinetics-400 & 28.6 \\\hline
         LSTC (ours) & I3D & Kinetics-400 & 22.4 \\
         LSTC (ours) & Res50 & Kinetics-400 & 28.7 \\
          LSTC (ours) & Res101-NL & Kinetics-600 & \textbf{30.3} \\\hline
    \end{tabular}
    \caption{Comparison results with other methods on AVA v2.2 validation set. The SlowFast results are reported in the official code repository~\cite{PySlowfast}.}
    \label{tab:ava_v22}
\end{table}

In Table~\ref{tab:ava_v21} and Table~\ref{tab:ava_v22}, we list the comparison results on the standard AVA benchmarks with other methods. It can be observed that on both version of AVA validation set, our method outperforms most of other methods. Especially, with similar backbone and pretraining settings, our LSTC can outperform the SlowFast~\cite{SlowFast}, LFB~\cite{LFB} and C-RCNN~\cite{wu2020context} counterpart with marginal computation cost (SlowFast processes 28.6 clips per second 
while ours achieves a processing rate of 27.5 clips in each second). It is also noticeable that the performance of our model with \emph{solely short-term context} (25.6 on v2.1 and 26.1 on v2.2) is still better than SlowFast~\cite{SlowFast}, AVSlowFast~\cite{xiao2020audiovisual} and comparable to LFB~\cite{LFB}. These comparison demonstrate that our LSTC is well suitable for atomic action detection. In Table~\ref{tab:HIEVE}, we compare our scheme with other methods on recently public HiEve datasets. It is observed that our LSTC outperforms other methods in both mAP and w-mAP with similar backbone network, indicating that our approach can properly handle person-level action detection in crowd surveillance scenes.

\begin{table}[]
    \centering
    \small
    \setlength{\belowcaptionskip}{-0.0cm}
    \begin{tabular}{c|c|cc}\hline
        method & backbone & mAP & w-mAP \\\hline
        RPN+I3D~\cite{HIEVE} & I3D & 8.3 & 6.8 \\
        VAT~\cite{girdhar2019actionTransformer} & I3D & 7.0 & 7.3 \\
        SlowFast~\cite{SlowFast} & Res50 & 7.4 & 5.3 \\\hline
        LSTC (ours)~ & Res50 & \textbf{8.9} & \textbf{7.4} \\\hline
    \end{tabular}
    \caption{Comparison results on HiEve test set. The values of mAP and w-mAP are averaged over all thresholds.}
    \label{tab:HIEVE}
\end{table}
\begin{figure*}
    \centering
    \includegraphics[width=0.90\textwidth]{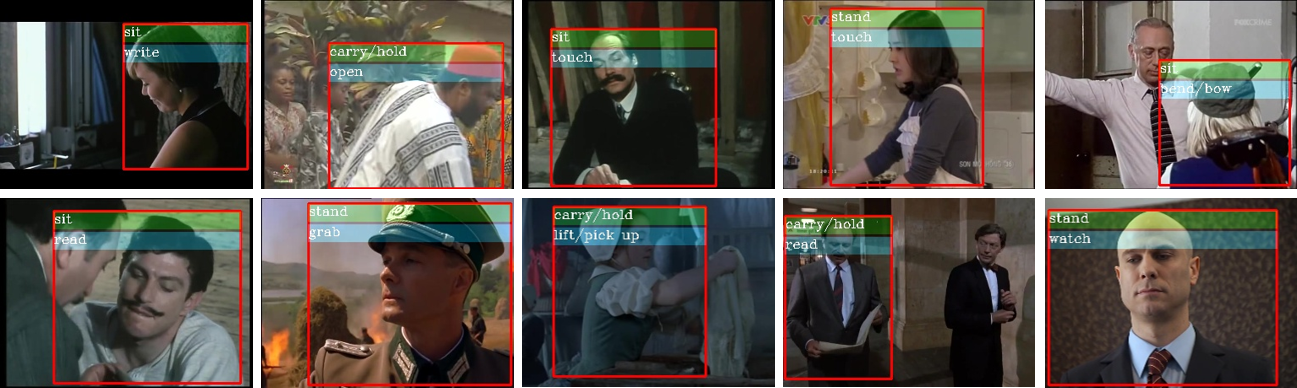}
    \caption{Qualitative results with most confident prediction via only short-term context (in green rectangle) and only long-term context (in blue rectangle, best viewed in color).}
    \label{fig:vis}
\end{figure*}

\subsection{Qualitative results}

In Figure~\ref{fig:vis}, we investigate the independent prediction results from either short-term or long-term context. To be specific, we only pass $\mathcal{Z}_s$ or $\mathcal{Z}_l$ to normalization function and generate final classification $\mathcal{C}$. We visualize the bounding boxes together with the most confident action type obtained from each temporal scope. From Figure~\ref{fig:vis}, we find the long-term and short-term context can provide complementary cues for prediction, thus the most confident action type from each branch can be different in most cases. The short-term context will guide the classifier to prefer pose or short interaction, while long-term context leans to some actions not captured completely in current shot but lasting for a relatively long time span, which can be inferred from the video segments before and after current timestamp. 

\section{Conclusion}
\label{Conclusion}
In this paper, we analyze the atomic action detection problem from the aspects of both long-term and short-term temporal span. We first build a context-based probability graph model and derive the conditional independence between two types of context. Guided by this conclusion, we decouple the task into two inference procedure with different spatiotemporal cues for atomic action recognition. For short-term context, we introduce dense attention to aggregate cues, for long-term information, second-order attention is designed to well handle long-term reliance. Results on challenging benchmarks demonstrate our LSTC is beneficial to high-quality action detection.  

\section{Acknowledgement}
This paper is supported in part by the following grants: National Key Research and Development Program of China Grant (No.2018AAA0\\100400), National Natural Science Foundation of China (No. 61971277).

\bibliographystyle{ACM-Reference-Format}
\bibliography{egbib}


\begin{thebibliography}{35}


\ifx \showCODEN    \undefined \def \showCODEN     #1{\unskip}     \fi
\ifx \showDOI      \undefined \def \showDOI       #1{#1}\fi
\ifx \showISBNx    \undefined \def \showISBNx     #1{\unskip}     \fi
\ifx \showISBNxiii \undefined \def \showISBNxiii  #1{\unskip}     \fi
\ifx \showISSN     \undefined \def \showISSN      #1{\unskip}     \fi
\ifx \showLCCN     \undefined \def \showLCCN      #1{\unskip}     \fi
\ifx \shownote     \undefined \def \shownote      #1{#1}          \fi
\ifx \showarticletitle \undefined \def \showarticletitle #1{#1}   \fi
\ifx \showURL      \undefined \def \showURL       {\relax}        \fi
\providecommand\bibfield[2]{#2}
\providecommand\bibinfo[2]{#2}
\providecommand\natexlab[1]{#1}
\providecommand\showeprint[2][]{arXiv:#2}

\bibitem[\protect\citeauthoryear{Bengio, Ducharme, Vincent, and Jauvin}{Bengio
  et~al\mbox{.}}{2006}]%
        {language_model}
\bibfield{author}{\bibinfo{person}{Y. Bengio}, \bibinfo{person}{Réjean
  Ducharme}, \bibinfo{person}{Pascal Vincent}, {and} \bibinfo{person}{Christian
  Jauvin}.} \bibinfo{year}{2006}\natexlab{}.
\newblock \bibinfo{booktitle}{\emph{Neural Probabilistic Language Models}}.
  Vol.~\bibinfo{volume}{3}.
\newblock \bibinfo{pages}{137--186}.
\newblock


\bibitem[\protect\citeauthoryear{Carreira and Zisserman}{Carreira and
  Zisserman}{2017}]%
        {I3D}
\bibfield{author}{\bibinfo{person}{Joao Carreira} {and} \bibinfo{person}{Andrew
  Zisserman}.} \bibinfo{year}{2017}\natexlab{}.
\newblock \showarticletitle{Quo vadis, action recognition? a new model and the
  kinetics dataset}. In \bibinfo{booktitle}{\emph{proceedings of the IEEE
  Conference on Computer Vision and Pattern Recognition (CVPR)}}.
  \bibinfo{pages}{6299--6308}.
\newblock


\bibitem[\protect\citeauthoryear{Deng, Dong, Socher, Li, Li, and Fei-Fei}{Deng
  et~al\mbox{.}}{2009}]%
        {ImageNet}
\bibfield{author}{\bibinfo{person}{Jia Deng}, \bibinfo{person}{Wei Dong},
  \bibinfo{person}{Richard Socher}, \bibinfo{person}{Li-Jia Li},
  \bibinfo{person}{Kai Li}, {and} \bibinfo{person}{Li Fei-Fei}.}
  \bibinfo{year}{2009}\natexlab{}.
\newblock \showarticletitle{Imagenet: A large-scale hierarchical image
  database}. In \bibinfo{booktitle}{\emph{Proc. IEEE Conference on Computer
  Vision and Pattern Recognition (CVPR)}}. \bibinfo{pages}{248--255}.
\newblock


\bibitem[\protect\citeauthoryear{Devlin, Chang, Lee, and Toutanova}{Devlin
  et~al\mbox{.}}{2019}]%
        {devlin-etal-2019-bert}
\bibfield{author}{\bibinfo{person}{Jacob Devlin}, \bibinfo{person}{Ming-Wei
  Chang}, \bibinfo{person}{Kenton Lee}, {and} \bibinfo{person}{Kristina
  Toutanova}.} \bibinfo{year}{2019}\natexlab{}.
\newblock \showarticletitle{{BERT}: Pre-training of Deep Bidirectional
  Transformers for Language Understanding}. In \bibinfo{booktitle}{\emph{ACL}}.
\newblock


\bibitem[\protect\citeauthoryear{Everingham, Eslami, Van~Gool, Williams, Winn,
  and Zisserman}{Everingham et~al\mbox{.}}{2015}]%
        {VOC}
\bibfield{author}{\bibinfo{person}{M. Everingham}, \bibinfo{person}{S.~M.~A.
  Eslami}, \bibinfo{person}{L. Van~Gool}, \bibinfo{person}{C.~K.~I. Williams},
  \bibinfo{person}{J. Winn}, {and} \bibinfo{person}{A. Zisserman}.}
  \bibinfo{year}{2015}\natexlab{}.
\newblock \showarticletitle{The Pascal Visual Object Classes Challenge: A
  Retrospective}.
\newblock \bibinfo{journal}{\emph{International Journal of Computer Vision
  (IJCV)}} \bibinfo{volume}{111}, \bibinfo{number}{1} (\bibinfo{date}{Jan.}
  \bibinfo{year}{2015}), \bibinfo{pages}{98--136}.
\newblock


\bibitem[\protect\citeauthoryear{Fan, Li, Xiong, Lo, and Feichtenhofer}{Fan
  et~al\mbox{.}}{2020}]%
        {PySlowfast}
\bibfield{author}{\bibinfo{person}{Haoqi Fan}, \bibinfo{person}{Yanghao Li},
  \bibinfo{person}{Bo Xiong}, \bibinfo{person}{Wan-Yen Lo}, {and}
  \bibinfo{person}{Christoph Feichtenhofer}.} \bibinfo{year}{2020}\natexlab{}.
\newblock \bibinfo{title}{PySlowFast}.
\newblock
  \bibinfo{howpublished}{\url{https://github.com/facebookresearch/slowfast}}.
\newblock


\bibitem[\protect\citeauthoryear{Feichtenhofer, Fan, Malik, and
  He}{Feichtenhofer et~al\mbox{.}}{2019}]%
        {SlowFast}
\bibfield{author}{\bibinfo{person}{Christoph Feichtenhofer},
  \bibinfo{person}{Haoqi Fan}, \bibinfo{person}{Jitendra Malik}, {and}
  \bibinfo{person}{Kaiming He}.} \bibinfo{year}{2019}\natexlab{}.
\newblock \showarticletitle{Slowfast networks for video recognition}. In
  \bibinfo{booktitle}{\emph{Proc. IEEE Conference on Computer Vision and
  Pattern Recognition (CVPR)}}. \bibinfo{pages}{6202--6211}.
\newblock


\bibitem[\protect\citeauthoryear{Girdhar, Carreira, Doersch, and
  Zisserman}{Girdhar et~al\mbox{.}}{2019}]%
        {girdhar2019actionTransformer}
\bibfield{author}{\bibinfo{person}{Rohit Girdhar}, \bibinfo{person}{Joao
  Carreira}, \bibinfo{person}{Carl Doersch}, {and} \bibinfo{person}{Andrew
  Zisserman}.} \bibinfo{year}{2019}\natexlab{}.
\newblock \showarticletitle{Video action transformer network}. In
  \bibinfo{booktitle}{\emph{Proc. IEEE Conference on Computer Vision and
  Pattern Recognition (CVPR)}}. \bibinfo{pages}{244--253}.
\newblock


\bibitem[\protect\citeauthoryear{Girdhar and Ramanan}{Girdhar and
  Ramanan}{2017}]%
        {girdhar2017attentionalPooling}
\bibfield{author}{\bibinfo{person}{Rohit Girdhar} {and} \bibinfo{person}{Deva
  Ramanan}.} \bibinfo{year}{2017}\natexlab{}.
\newblock \showarticletitle{Attentional pooling for action recognition}. In
  \bibinfo{booktitle}{\emph{Proc. Advances in Neural Information Processing
  Systems (NeurIPS)}}. \bibinfo{pages}{34--45}.
\newblock


\bibitem[\protect\citeauthoryear{Gu, Sun, Ross, Vondrick, Pantofaru, Li,
  Vijayanarasimhan, Toderici, Ricco, Sukthankar, et~al\mbox{.}}{Gu
  et~al\mbox{.}}{2018}]%
        {AVA}
\bibfield{author}{\bibinfo{person}{Chunhui Gu}, \bibinfo{person}{Chen Sun},
  \bibinfo{person}{David~A Ross}, \bibinfo{person}{Carl Vondrick},
  \bibinfo{person}{Caroline Pantofaru}, \bibinfo{person}{Yeqing Li},
  \bibinfo{person}{Sudheendra Vijayanarasimhan}, \bibinfo{person}{George
  Toderici}, \bibinfo{person}{Susanna Ricco}, \bibinfo{person}{Rahul
  Sukthankar}, {et~al\mbox{.}}} \bibinfo{year}{2018}\natexlab{}.
\newblock \showarticletitle{Ava: A video dataset of spatio-temporally localized
  atomic visual actions}. In \bibinfo{booktitle}{\emph{Proc. IEEE Conference on
  Computer Vision and Pattern Recognition (CVPR)}}.
  \bibinfo{pages}{6047--6056}.
\newblock


\bibitem[\protect\citeauthoryear{Guo, Li, Li, and Lin}{Guo
  et~al\mbox{.}}{2018}]%
        {guo2018nd}
\bibfield{author}{\bibinfo{person}{Jianbo Guo}, \bibinfo{person}{Yuxi Li},
  \bibinfo{person}{Jianguo Li}, {and} \bibinfo{person}{Weiyao Lin}.}
  \bibinfo{year}{2018}\natexlab{}.
\newblock \showarticletitle{{Network Decoupling}: From Regular Convolution to
  Separable Depthwise Convolution}. In \bibinfo{booktitle}{\emph{BMVC}}.
\newblock


\bibitem[\protect\citeauthoryear{He, Gkioxari, Dollar, and Girshick}{He
  et~al\mbox{.}}{2017}]%
        {He_2017_ICCV}
\bibfield{author}{\bibinfo{person}{Kaiming He}, \bibinfo{person}{Georgia
  Gkioxari}, \bibinfo{person}{Piotr Dollar}, {and} \bibinfo{person}{Ross
  Girshick}.} \bibinfo{year}{2017}\natexlab{}.
\newblock \showarticletitle{Mask R-CNN}. In
  \bibinfo{booktitle}{\emph{Proceedings of the IEEE International Conference on
  Computer Vision (ICCV)}}.
\newblock


\bibitem[\protect\citeauthoryear{He, Zhang, Ren, and Sun}{He
  et~al\mbox{.}}{2016}]%
        {he2016deep}
\bibfield{author}{\bibinfo{person}{Kaiming He}, \bibinfo{person}{Xiangyu
  Zhang}, \bibinfo{person}{Shaoqing Ren}, {and} \bibinfo{person}{Jian Sun}.}
  \bibinfo{year}{2016}\natexlab{}.
\newblock \showarticletitle{Deep residual learning for image recognition}. In
  \bibinfo{booktitle}{\emph{Proceedings of the IEEE conference on computer
  vision and pattern recognition (CVPR)}}. \bibinfo{pages}{770--778}.
\newblock


\bibitem[\protect\citeauthoryear{Lin, Doll{\'a}r, Girshick, He, Hariharan, and
  Belongie}{Lin et~al\mbox{.}}{2017}]%
        {lin2017feature}
\bibfield{author}{\bibinfo{person}{Tsung-Yi Lin}, \bibinfo{person}{Piotr
  Doll{\'a}r}, \bibinfo{person}{Ross Girshick}, \bibinfo{person}{Kaiming He},
  \bibinfo{person}{Bharath Hariharan}, {and} \bibinfo{person}{Serge Belongie}.}
  \bibinfo{year}{2017}\natexlab{}.
\newblock \showarticletitle{Feature pyramid networks for object detection}. In
  \bibinfo{booktitle}{\emph{Proceedings of the IEEE conference on computer
  vision and pattern recognition (CVPR)}}. \bibinfo{pages}{2117--2125}.
\newblock


\bibitem[\protect\citeauthoryear{Lin, Maire, Belongie, Hays, Perona, Ramanan,
  Doll{\'a}r, and Zitnick}{Lin et~al\mbox{.}}{2014}]%
        {COCO}
\bibfield{author}{\bibinfo{person}{Tsung-Yi Lin}, \bibinfo{person}{Michael
  Maire}, \bibinfo{person}{Serge Belongie}, \bibinfo{person}{James Hays},
  \bibinfo{person}{Pietro Perona}, \bibinfo{person}{Deva Ramanan},
  \bibinfo{person}{Piotr Doll{\'a}r}, {and} \bibinfo{person}{C~Lawrence
  Zitnick}.} \bibinfo{year}{2014}\natexlab{}.
\newblock \showarticletitle{Microsoft coco: Common objects in context}. In
  \bibinfo{booktitle}{\emph{European conference on computer vision (ECCV)}}.
  Springer, \bibinfo{pages}{740--755}.
\newblock


\bibitem[\protect\citeauthoryear{Lin, Liu, Liu, Li, Qi, Qian, Wang, Sebe, Xu,
  Xiong, and Shah}{Lin et~al\mbox{.}}{2020}]%
        {HIEVE}
\bibfield{author}{\bibinfo{person}{Weiyao Lin}, \bibinfo{person}{Huabin Liu},
  \bibinfo{person}{Shizhan Liu}, \bibinfo{person}{Yuxi Li},
  \bibinfo{person}{Guo-Jun Qi}, \bibinfo{person}{Rui Qian},
  \bibinfo{person}{Tao Wang}, \bibinfo{person}{Nicu Sebe},
  \bibinfo{person}{Ning Xu}, \bibinfo{person}{Hongkai Xiong}, {and}
  \bibinfo{person}{Mubarak Shah}.} \bibinfo{year}{2020}\natexlab{}.
\newblock \bibinfo{title}{Human in Events: A Large-Scale Benchmark for
  Human-centric Video Analysis in Complex Events}.
\newblock
\newblock
\showeprint[arxiv]{2005.04490}~[cs.CV]


\bibitem[\protect\citeauthoryear{Marszalek, Laptev, and Schmid}{Marszalek
  et~al\mbox{.}}{2009}]%
        {marszalek2009actionsInContext}
\bibfield{author}{\bibinfo{person}{Marcin Marszalek}, \bibinfo{person}{Ivan
  Laptev}, {and} \bibinfo{person}{Cordelia Schmid}.}
  \bibinfo{year}{2009}\natexlab{}.
\newblock \showarticletitle{Actions in context}. In
  \bibinfo{booktitle}{\emph{Proc. IEEE Conference on Computer Vision and
  Pattern Recognition (CVPR)}}. IEEE, \bibinfo{pages}{2929--2936}.
\newblock


\bibitem[\protect\citeauthoryear{{Oh}, {Lee}, {Xu}, and {Kim}}{{Oh}
  et~al\mbox{.}}{2020}]%
        {STM}
\bibfield{author}{\bibinfo{person}{S.~W. {Oh}}, \bibinfo{person}{J.~Y. {Lee}},
  \bibinfo{person}{N. {Xu}}, {and} \bibinfo{person}{S.~J. {Kim}}.}
  \bibinfo{year}{2020}\natexlab{}.
\newblock \showarticletitle{Space-time Memory Networks for Video Object
  Segmentation with User Guidance}.
\newblock \bibinfo{journal}{\emph{IEEE Transactions on Pattern Analysis and
  Machine Intelligence (PAMI)}} (\bibinfo{year}{2020}), \bibinfo{pages}{1--1}.
\newblock


\bibitem[\protect\citeauthoryear{Pennington, Socher, and Manning}{Pennington
  et~al\mbox{.}}{2014}]%
        {pennington2014glove}
\bibfield{author}{\bibinfo{person}{Jeffrey Pennington},
  \bibinfo{person}{Richard Socher}, {and} \bibinfo{person}{Christopher~D
  Manning}.} \bibinfo{year}{2014}\natexlab{}.
\newblock \showarticletitle{Glove: Global vectors for word representation}. In
  \bibinfo{booktitle}{\emph{Proceedings of the 2014 conference on empirical
  methods in natural language processing (EMNLP)}}.
  \bibinfo{pages}{1532--1543}.
\newblock


\bibitem[\protect\citeauthoryear{Ren, He, Girshick, and Sun}{Ren
  et~al\mbox{.}}{2015}]%
        {FasterRCNN}
\bibfield{author}{\bibinfo{person}{Shaoqing Ren}, \bibinfo{person}{Kaiming He},
  \bibinfo{person}{Ross Girshick}, {and} \bibinfo{person}{Jian Sun}.}
  \bibinfo{year}{2015}\natexlab{}.
\newblock \showarticletitle{Faster r-cnn: Towards real-time object detection
  with region proposal networks}. In \bibinfo{booktitle}{\emph{Proc. Advances
  in Neural Information Processing Systems (NeurIPS)}}.
  \bibinfo{pages}{91--99}.
\newblock


\bibitem[\protect\citeauthoryear{Sharma, Kiros, and Salakhutdinov}{Sharma
  et~al\mbox{.}}{2015}]%
        {sharma2015actionVisualAtt}
\bibfield{author}{\bibinfo{person}{Shikhar Sharma}, \bibinfo{person}{Ryan
  Kiros}, {and} \bibinfo{person}{Ruslan Salakhutdinov}.}
  \bibinfo{year}{2015}\natexlab{}.
\newblock \showarticletitle{Action recognition using visual attention}.
\newblock \bibinfo{journal}{\emph{arXiv preprint arXiv:1511.04119}}
  (\bibinfo{year}{2015}).
\newblock


\bibitem[\protect\citeauthoryear{Simonyan and Zisserman}{Simonyan and
  Zisserman}{2014}]%
        {TwoStream}
\bibfield{author}{\bibinfo{person}{Karen Simonyan} {and}
  \bibinfo{person}{Andrew Zisserman}.} \bibinfo{year}{2014}\natexlab{}.
\newblock \showarticletitle{Two-stream convolutional networks for action
  recognition in videos}. In \bibinfo{booktitle}{\emph{Proc. Advances in Neural
  Information Processing Systems (NeurIPS)}}. \bibinfo{pages}{568--576}.
\newblock


\bibitem[\protect\citeauthoryear{Soomro, Zamir, and Shah}{Soomro
  et~al\mbox{.}}{2012}]%
        {UCF101}
\bibfield{author}{\bibinfo{person}{Khurram Soomro},
  \bibinfo{person}{Amir~Roshan Zamir}, {and} \bibinfo{person}{Mubarak Shah}.}
  \bibinfo{year}{2012}\natexlab{}.
\newblock \showarticletitle{UCF101: A dataset of 101 human actions classes from
  videos in the wild}.
\newblock \bibinfo{journal}{\emph{arXiv preprint arXiv:1212.0402}}
  (\bibinfo{year}{2012}).
\newblock


\bibitem[\protect\citeauthoryear{Sun, Shrivastava, Vondrick, Murphy,
  Sukthankar, and Schmid}{Sun et~al\mbox{.}}{2018}]%
        {ACRN}
\bibfield{author}{\bibinfo{person}{Chen Sun}, \bibinfo{person}{Abhinav
  Shrivastava}, \bibinfo{person}{Carl Vondrick}, \bibinfo{person}{Kevin
  Murphy}, \bibinfo{person}{Rahul Sukthankar}, {and} \bibinfo{person}{Cordelia
  Schmid}.} \bibinfo{year}{2018}\natexlab{}.
\newblock \showarticletitle{Actor-centric relation network}. In
  \bibinfo{booktitle}{\emph{Proc. European Conference on Computer Vision
  (ECCV)}}. \bibinfo{pages}{318--334}.
\newblock


\bibitem[\protect\citeauthoryear{Tran, Bourdev, Fergus, Torresani, and
  Paluri}{Tran et~al\mbox{.}}{2015}]%
        {C3D}
\bibfield{author}{\bibinfo{person}{Du Tran}, \bibinfo{person}{Lubomir Bourdev},
  \bibinfo{person}{Rob Fergus}, \bibinfo{person}{Lorenzo Torresani}, {and}
  \bibinfo{person}{Manohar Paluri}.} \bibinfo{year}{2015}\natexlab{}.
\newblock \showarticletitle{Learning spatiotemporal features with 3d
  convolutional networks}. In \bibinfo{booktitle}{\emph{Proc. IEEE
  International Conference on Computer Vision (ICCV)}}.
  \bibinfo{pages}{4489--4497}.
\newblock


\bibitem[\protect\citeauthoryear{Vaswani, Shazeer, Parmar, Uszkoreit, Jones,
  Gomez, Kaiser, and Polosukhin}{Vaswani et~al\mbox{.}}{2017}]%
        {vaswani2017AttentionAllYouNeed}
\bibfield{author}{\bibinfo{person}{Ashish Vaswani}, \bibinfo{person}{Noam
  Shazeer}, \bibinfo{person}{Niki Parmar}, \bibinfo{person}{Jakob Uszkoreit},
  \bibinfo{person}{Llion Jones}, \bibinfo{person}{Aidan~N Gomez},
  \bibinfo{person}{{\L}ukasz Kaiser}, {and} \bibinfo{person}{Illia
  Polosukhin}.} \bibinfo{year}{2017}\natexlab{}.
\newblock \showarticletitle{Attention is all you need}. In
  \bibinfo{booktitle}{\emph{Proc. Advances in Neural Information Processing
  Systems (NeurIPS)}}. \bibinfo{pages}{5998--6008}.
\newblock


\bibitem[\protect\citeauthoryear{Wang, Tran, Torresani, and Feiszli}{Wang
  et~al\mbox{.}}{2020}]%
        {Wang_2020_CVPR}
\bibfield{author}{\bibinfo{person}{Heng Wang}, \bibinfo{person}{Du Tran},
  \bibinfo{person}{Lorenzo Torresani}, {and} \bibinfo{person}{Matt Feiszli}.}
  \bibinfo{year}{2020}\natexlab{}.
\newblock \showarticletitle{Video Modeling With Correlation Networks}. In
  \bibinfo{booktitle}{\emph{IEEE/CVF Conference on Computer Vision and Pattern
  Recognition (CVPR)}}.
\newblock


\bibitem[\protect\citeauthoryear{Wang, Xiong, Wang, Qiao, Lin, Tang, and
  Van~Gool}{Wang et~al\mbox{.}}{2016}]%
        {TSN}
\bibfield{author}{\bibinfo{person}{Limin Wang}, \bibinfo{person}{Yuanjun
  Xiong}, \bibinfo{person}{Zhe Wang}, \bibinfo{person}{Yu Qiao},
  \bibinfo{person}{Dahua Lin}, \bibinfo{person}{Xiaoou Tang}, {and}
  \bibinfo{person}{Luc Van~Gool}.} \bibinfo{year}{2016}\natexlab{}.
\newblock \showarticletitle{Temporal segment networks: Towards good practices
  for deep action recognition}. In \bibinfo{booktitle}{\emph{Proc. European
  Conference on Computer Vision (ECCV)}}. Springer, \bibinfo{pages}{20--36}.
\newblock


\bibitem[\protect\citeauthoryear{Wang, Girshick, Gupta, and He}{Wang
  et~al\mbox{.}}{2018}]%
        {Nonlocal}
\bibfield{author}{\bibinfo{person}{Xiaolong Wang}, \bibinfo{person}{Ross
  Girshick}, \bibinfo{person}{Abhinav Gupta}, {and} \bibinfo{person}{Kaiming
  He}.} \bibinfo{year}{2018}\natexlab{}.
\newblock \showarticletitle{Non-local neural networks}. In
  \bibinfo{booktitle}{\emph{Proc. IEEE Conference on Computer Vision and
  Pattern Recognition (CVPR)}}. \bibinfo{pages}{7794--7803}.
\newblock


\bibitem[\protect\citeauthoryear{Wu, Feichtenhofer, Fan, He, Krahenbuhl, and
  Girshick}{Wu et~al\mbox{.}}{2019}]%
        {LFB}
\bibfield{author}{\bibinfo{person}{Chao-Yuan Wu}, \bibinfo{person}{Christoph
  Feichtenhofer}, \bibinfo{person}{Haoqi Fan}, \bibinfo{person}{Kaiming He},
  \bibinfo{person}{Philipp Krahenbuhl}, {and} \bibinfo{person}{Ross Girshick}.}
  \bibinfo{year}{2019}\natexlab{}.
\newblock \showarticletitle{Long-term feature banks for detailed video
  understanding}. In \bibinfo{booktitle}{\emph{Proc. IEEE Conference on
  Computer Vision and Pattern Recognition (CVPR)}}. \bibinfo{pages}{284--293}.
\newblock


\bibitem[\protect\citeauthoryear{Wu, Kuang, Wang, Zhang, and Wu}{Wu
  et~al\mbox{.}}{2020}]%
        {wu2020context}
\bibfield{author}{\bibinfo{person}{Jianchao Wu}, \bibinfo{person}{Zhanghui
  Kuang}, \bibinfo{person}{Limin Wang}, \bibinfo{person}{Wayne Zhang}, {and}
  \bibinfo{person}{Gangshan Wu}.} \bibinfo{year}{2020}\natexlab{}.
\newblock \showarticletitle{Context-aware rcnn: A baseline for action detection
  in videos}. In \bibinfo{booktitle}{\emph{European Conference on Computer
  Vision (ECCV)}}. Springer, \bibinfo{pages}{440--456}.
\newblock


\bibitem[\protect\citeauthoryear{Xiao, Lee, Grauman, Malik, and
  Feichtenhofer}{Xiao et~al\mbox{.}}{2020}]%
        {xiao2020audiovisual}
\bibfield{author}{\bibinfo{person}{Fanyi Xiao}, \bibinfo{person}{Yong~Jae Lee},
  \bibinfo{person}{Kristen Grauman}, \bibinfo{person}{Jitendra Malik}, {and}
  \bibinfo{person}{Christoph Feichtenhofer}.} \bibinfo{year}{2020}\natexlab{}.
\newblock \bibinfo{title}{Audiovisual SlowFast Networks for Video Recognition}.
\newblock
\newblock
\showeprint[arxiv]{2001.08740}~[cs.CV]


\bibitem[\protect\citeauthoryear{Xie, Sun, Huang, Tu, and Murphy}{Xie
  et~al\mbox{.}}{2018}]%
        {S3D}
\bibfield{author}{\bibinfo{person}{Saining Xie}, \bibinfo{person}{Chen Sun},
  \bibinfo{person}{Jonathan Huang}, \bibinfo{person}{Zhuowen Tu}, {and}
  \bibinfo{person}{Kevin Murphy}.} \bibinfo{year}{2018}\natexlab{}.
\newblock \showarticletitle{Rethinking spatiotemporal feature learning:
  Speed-accuracy trade-offs in video classification}. In
  \bibinfo{booktitle}{\emph{Proc. European Conference on Computer Vision
  (ECCV)}}. \bibinfo{pages}{305--321}.
\newblock


\bibitem[\protect\citeauthoryear{Yao and Fei-Fei}{Yao and Fei-Fei}{2010}]%
        {yao2010modeling}
\bibfield{author}{\bibinfo{person}{Bangpeng Yao} {and} \bibinfo{person}{Li
  Fei-Fei}.} \bibinfo{year}{2010}\natexlab{}.
\newblock \showarticletitle{Modeling mutual context of object and human pose in
  human-object interaction activities}. In \bibinfo{booktitle}{\emph{Proc. IEEE
  Conference on Computer Vision and Pattern Recognition (CVPR)}}. IEEE,
  \bibinfo{pages}{17--24}.
\newblock


\bibitem[\protect\citeauthoryear{Zhang, Tokmakov, Hebert, and Schmid}{Zhang
  et~al\mbox{.}}{2019}]%
        {zhang2019Graph}
\bibfield{author}{\bibinfo{person}{Yubo Zhang}, \bibinfo{person}{Pavel
  Tokmakov}, \bibinfo{person}{Martial Hebert}, {and} \bibinfo{person}{Cordelia
  Schmid}.} \bibinfo{year}{2019}\natexlab{}.
\newblock \showarticletitle{A structured model for action detection}. In
  \bibinfo{booktitle}{\emph{Proc. IEEE Conference on Computer Vision and
  Pattern Recognition (CVPR)}}. \bibinfo{pages}{9975--9984}.
\newblock


\end{thebibliography}










\end{document}